%% file: main.tex
\documentclass[lettersize,journal]{IEEEtran}
\usepackage{amsmath,amsfonts,amssymb}
\usepackage{algorithmic}
\usepackage{algorithm}
\usepackage{array}
\usepackage[caption=false,font=normalsize,labelfont=sf,textfont=sf]{subfig}
\usepackage{textcomp}
\usepackage{stfloats}
\usepackage{url}
\usepackage{verbatim}
\usepackage{graphicx}
\usepackage{multirow} 
\def\BibTeX{{\rm B\kern-.05em{\sc i\kern-.025em b}\kern-.08em
    T\kern-.1667em\lower.7ex\hbox{E}\kern-.125emX}}
\usepackage{balance}

\usepackage{hyperref}
\usepackage{url}

\usepackage{cuted}
\usepackage{wrapfig}
\usepackage{booktabs}
\usepackage{soul}
\usepackage[table]{xcolor}
\usepackage{pifont}
\usepackage{etoolbox}
\usepackage{siunitx}
\usepackage{enumitem,kantlipsum}
\usepackage{capt-of} 
\newcommand{\ms}[2]{$\num{#1} \pm {\scriptstyle\num{#2}}$}
\definecolor{lightgreen}{RGB}{70, 240, 94}
\definecolor{darkgreen}{RGB}{33, 181, 52}
\definecolor{lightcyan}{RGB}{204, 255, 255}
\definecolor{darkblue}{RGB}{25, 198, 212}
\definecolor{lightred}{RGB}{250, 92, 118}

\makeatletter
\let\@oldmaketitle\@maketitle
\renewcommand{\@maketitle}{\@oldmaketitle

  \vspace{8pt}
  
  \begin{center}
    \includegraphics[width=1.0\textwidth]{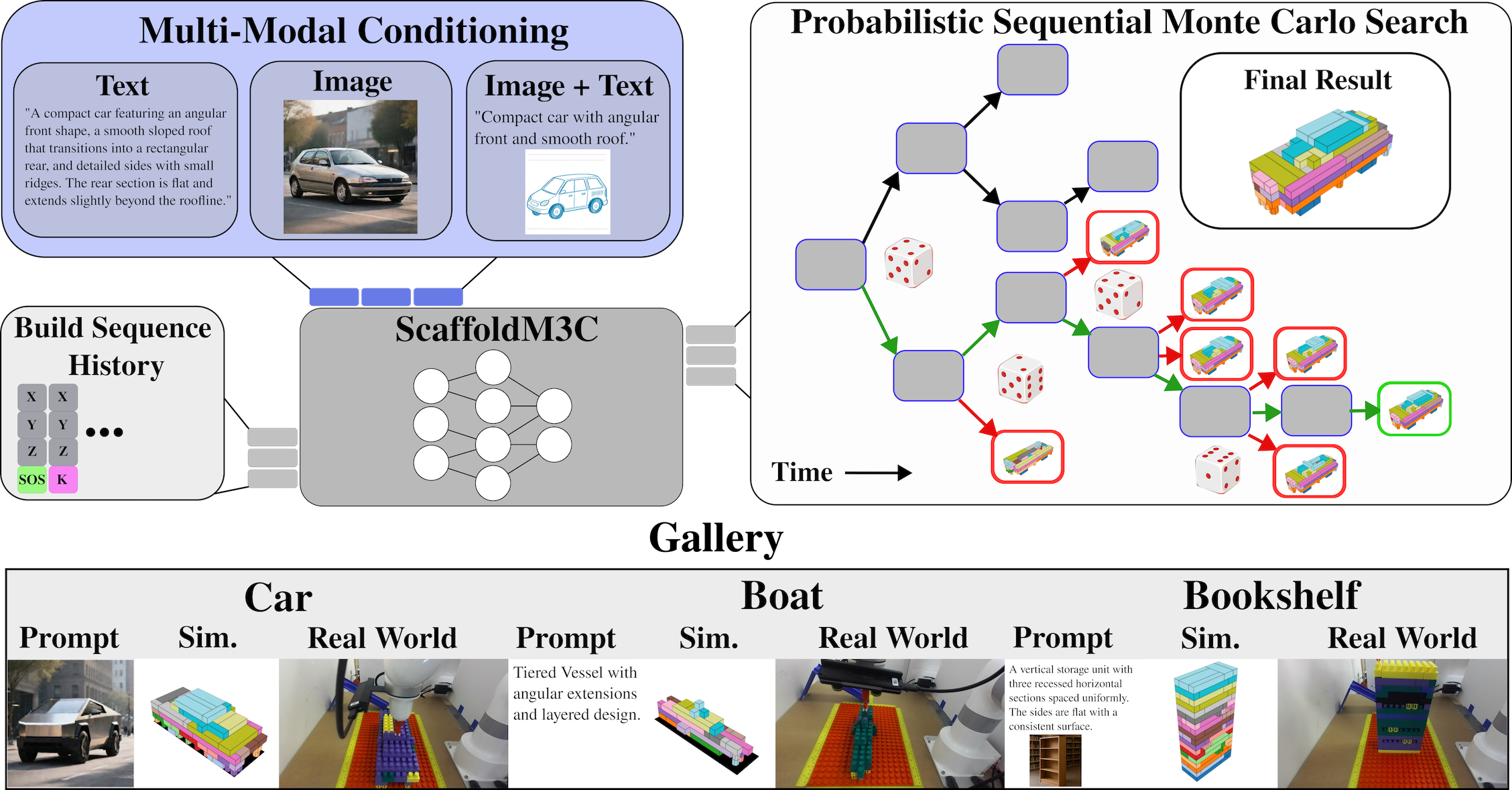}
    \setcounter{figure}{0}
    \captionof{figure}{We present ScaffoldM3C, a lightweight, open source 225M parameter Vision-Language-Construction (VLC) model, trained on 553K different conditioned assembly sequences. Given a starting build location, ScaffoldM3C auto-regressively generates a condition aligned step-by-step assembly plan. At each step, ScaffoldM3C proposes a set of next step candidate blocks from which to choose, while multiple possible assembly sequences are considered simultaneously through a sequential Monte Carlo algorithm, populated based on the candidate block likelihoods. We demonstrate the success of our architecture through simulated and robotic real world construction.}
    \label{fig:Cover_Fig}
  \end{center}
  \vspace{-20pt} 
}
\makeatother

\begin{document}
\title{\includegraphics[height=1.8em]{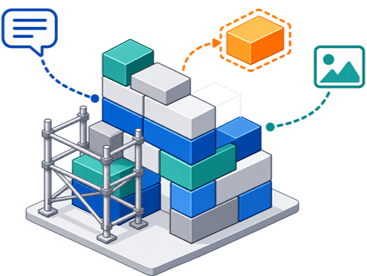}ScaffoldM3C: A Multimodal Sequential Monte \\ Carlo Framework for Generative Stable Construction Planning \\ 
{\large Project Page: \url{https://stanfordmsl.github.io/ScaffoldM3C/}}}
\author{Gadiel Sznaier Camps$^{1}$, Chengyang He$^{2}$, Guillaume Sartoretti$^{2}$, Eduardo Montijano$^{3}$, Mac Schwager$^{1}$
\thanks{$^1$ Multi-robot Systems Lab (MSL), Stanford University, Stanford, CA, USA}
\thanks{$^2$ Multi-Agent Robotic Motion Lab (MARMot), National University of Singapore, Singapore}
\thanks{$^3$ Perception-Oriented Control Team (POC), Universidad de Zaragoza, Zaragoza, Spain}
}

\maketitle

\begin{abstract}
Autonomously constructing physically realizable 3D structures remains a significant challenge due to combinatorial action spaces, interchangeable components, equifinal assembly sequences, and strict stability requirements during construction. State-of-the-art methods fine-tune large language models for text-based generative construction. However, these approaches do not allow for Multimodal (text, image, sketch) conditioning, overlook the practical role of scaffolding for stabilizing intermediate structures, and suffer from slow inference speeds. Therefore, we formulate construction as a probabilistic next-block generation task with multiple potential assembly actions and multiple potential task conditioning modalities. Concurrently, we explicitly consider the utility of scaffolding by introducing an auxiliary scaffold block token.
We present \textit{Scaffold Multimodal Monte Carlo (ScaffoldM3C)}, a multimodal, lightweight, auto-regressive model for stable block-based construction, that proposes a set of next-step candidate blocks. Leveraging these candidates, we utilize Sequential Monte Carlo (SMC) to maintain a population of possible assembly sequences, allowing us to consider multiple, potentially different, assembly directions simultaneously. We train our multimodal architecture by extending the StableText2Brick dataset \cite{pun2025generating} to contain image conditioning prompts and scaffold-stabilized build sequences.
ScaffoldM3C is $4\times$ smaller than competing baselines, yielding a $5\times$ to $20\times$ speedup during inference, while achieving comparable construction quality to state-of-the-art methods and higher overall stability. We demonstrate the effectiveness of our approach through simulations and real-world robot assembly demonstrations.

\end{abstract}


\section{Introduction}

Automated construction requires sophisticated long-horizon spatial, temporal, and semantic reasoning. Unlike other object generation problems, construction depends on physical assembly feasibility. Subassemblies often require temporary support until completion, and multiple component choices or sequence orders can yield the same final structure. Valid plans must therefore select components, determine placements, and order steps to ensure every intermediate state remains physically realizable and unobstructed.

Existing approaches simplify this problem by imposing strong assumptions on representation~\cite{zhang20223d}, conditioning modality~\cite{du2024generative}, or assembly processes~\cite{walsman2024learning}. Construction is typically formulated as a sequential decision-making over voxel spaces~\cite{earle2024dreamcraft} or with restricted libraries~\cite{yu2026brickcraft} to improve tractability at the cost of geometric fidelity and modeling placement uncertainty. Rule-based planning~\cite{joglekar2024towards}, search~\cite{tian2024asap}, or post-processing~\cite{pun2025generating} have been proposed to enforce physical validity post-generation. While resulting in improved stability, these strategies decouple generation from physical reasoning, requiring costly validity checks, rollbacks, or rigid constraints at inference~\cite{goldberg2025blox}.

Recent foundation-model approaches frame construction as autoregressive generation, adapting large language models for next-brick prediction from text prompts~\cite{pun2025generating}. However, this paradigm suffers from three fundamental limitations. First, text-only conditioning prevents users from providing visual or geometric guidance. Second, a lack of explicit reasoning over temporary support forces a heavy reliance on costly physics-based ``predict-evaluate-rollback'' pipelines. Third, current autoregressive generation methods rely on single-hypothesis rollouts, which discount the possibility of multiple valid actions, thus often rejecting more feasible long-horizon plans in favor of immediate short-term alignment.
To address these challenges, we propose \textbf{ScaffoldM3C}, a scaffolding-aware Vision-Language-Construction (VLC) framework for stable block-based assembly. Our main contributions are:

\noindent\textbf{(1) Probabilistic Multi-Hypothesis Assembly Formulation:}  We model assembly probabilistically by generating candidate block proposals with placement likelihoods, enabling Sequential Monte Carlo (SMC) inference to maintain diverse, feasible build sequences. We augment this search mechanism with specialized auxiliary scaffolding blocks to resolve intermediate stability issues during generation.

\noindent\textbf{(2) Lightweight Multimodal Task-Specific Architecture:} We introduce a dual-stream Vision-Language-Construction (VLC) architecture with interwoven cross-attention that decouples conditioning (text and images) from assembly, as well as an extended multimodal scaffolded dataset. 

Trained from scratch without relying on foundation models, \textbf{ScaffoldM3C} uses $4\times$ fewer parameters and achieves $5\times$ to $20\times$ faster inference while delivering comparable or superior overall structural stability.

\section{Related Work}

\noindent\textbf{Generative Autoregressive Construction:} Generative construction methods typically leverage large language models to propose sequential build steps from text instructions~\cite{joglekar2024towards, goldberg2025blox, pun2025generating}. To build complex structures, existing approaches rely on baseline spatial reasoning~\cite{joglekar2024towards, goldberg2025blox}, incorporate external physics simulators for online verification~\cite{goldberg2025blox}, finetune on simulation-verified~\cite{liu2023simulation, wen2026bricksim} stable build sequences~\cite{pun2025generating}, or expand component vocabularies for specific domains~\cite{ge2025lego, xu2025legoace, kulits2026bricknet}. Crucially, these methods deploy single-hypothesis rollouts at inference, making them vulnerable to short-horizon greedy placement errors that require expensive simulator rollbacks. Moreover, they lack temporary support reasoning, restricting the range of possible architectural geometries. In contrast, our model integrates multimodal inputs (text and images), generates explicit auxiliary scaffolding, and employs Sequential Monte Carlo (SMC) inference to track multiple candidate assembly sequences concurrently without rollbacks.

\noindent\textbf{Geometric Representations and Conditioning:} Beyond text conditioning, visual and spatial representations have also been explored for generative assembly. Image-conditioned assembly transformers leverage rich visual priors~\cite{guo2024treesba, yu2026brickcraft}, but suffer from 2D-3D correspondence ambiguity and component occlusions~\cite{yan2025neural, walsman2024learning}. Point-cloud models generate high-fidelity part placements~\cite{qi2017pointnet, xu2024spaformer, du2024generative, zhang20223d} through latent geometry encodings, yet combinatorial scaling restricts them to structures with few components. Voxel-based architectures operate on discretized, fixed-size spatial grids, simplifying the learning task and allowing straightforward occupancy comparisons~\cite{zhang2021learning, sudhakaran2021growing, earle2024dreamcraft, chen2026resembrick}. However, they neglect physical stability and suffer from cubic spatial scaling, limiting maximum structure size~\cite{zhang2021learning, ren2024xcube}. Rather than relying on rigid voxel grids or dense point clouds, our model directly outputs block coordinates in continuous space while supporting flexible multimodal conditioning.

\noindent\textbf{Stability-Aware Construction:} Ensuring physical stability during build sequences is a critical requirement in automated assembly. Prior research addresses structural integrity through analytical force-balancing optimizations~\cite{liu2024stablelego, yang2026far} or simulation of rigid-body and snap-fit mechanics~\cite{wen2026bricksim}. To filter out unstable placements, existing frameworks employ post-processing~\cite{liu2025physics, ota2024autonomous, pun2025generating}, policy optimization~\cite{xu2026rollback}, or a combination of both~\cite{ni2026brickanything, yuan2026sample}. However, these methods focus primarily on rejecting invalid actions or verifying final shapes, without considering intermediate support structures. In contrast, we adopt a simplified, conservative stability formulation specifically designed to highlight and demonstrate the utility of scaffolding during assembly.

\begin{figure*}[t]
    \centering
    \includegraphics[width=\textwidth]{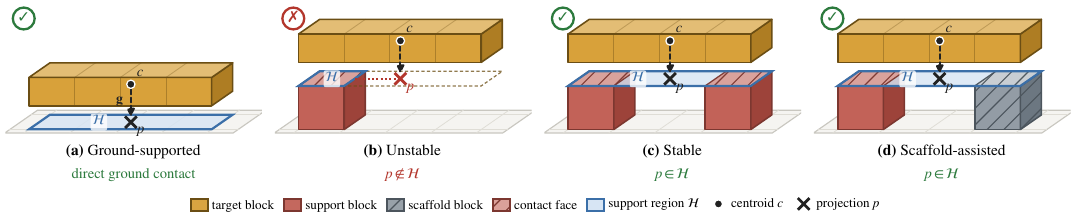}
    \caption{Stability constraints for candidate block placements. Each panel
  shows a target block being evaluated before placement: the centroid
  $c$ is projected vertically onto the contact plane at $p$, and the
  support region $\mathcal{H}$ (blue) is the 2D convex hull of the
  contact faces (hatched) provided by the blocks underneath.
  (a)~A block resting on the ground is stable by direct ground contact.
  (b)~With a single support, $p$ falls outside $\mathcal{H}$ and the
  placement is considered unstable.
  (c)~With supports at both ends, $\mathcal{H}$ spans the gap between
  their contact faces, so $p \in \mathcal{H}$ and the placement is
  stable.
  (d)~An auxiliary scaffold block (gray) restores the missing support,
  enlarging $\mathcal{H}$ so the otherwise-unstable placement becomes
  feasible.}
    \label{fig:stability_examples}
\end{figure*}

\section{Probabilistic Multi-Hypothesis Assembly Formulation}
\subsection{Stable Assembly Sequences}\label{sec:problem_formulation}
We model structure construction from a finite library of
block primitives (e.g., Lego or Tetris-style ``bricks''). Each block $b=(P,\kappa)$ consists of a 3D position $P=(x,y,z)\in\mathbb{R}^3$ and a shape type $\kappa\in\{1,\ldots,K\}$. Given a high-level specification $q$ (text, image, or multimodal), our objective is to generate an ordered sequence of $N_b$ blocks,
\begin{equation}
    \mathcal{S}=\langle b_1,b_2,\ldots,b_{N_b}\rangle,
\end{equation}
that forms the target structure while satisfying physical validity constraints.

A sequence $\mathcal{S}$ is valid if:  (i) the first block $b_1$ contacts the ground; (ii) no newly placed block overlaps existing ones; and (iii) every block is stable upon placement. A block is considered stable if it rests directly on the ground
or if the vertical projection $p$ of its centroid $c$ lies within
the 2D convex hull $\mathcal{H}$ of the contact faces
of the blocks directly underneath it, as illustrated in
Figure~\ref{fig:stability_examples}.

To support otherwise unstable blocks, we augment the library with an auxiliary scaffold block featuring a $1\times1$ cross-section and adjustable height. Scaffolds bridge gaps to the ground or blocks underneath and must be placed prior to the target block.

\subsection{Assembly Distributions}
At step $t$ in an assembly task, zero or more candidate block placements may be feasible under stability, collision, and precedence constraints. Therefore, we model the next construction step as a distribution over candidate blocks $\mathcal{U}_t$, comprising standard and scaffold blocks. The candidate set is defined as
\begin{equation}
        \mathcal{C}_t = \left\{ \left(b_j^{(t)},\rho_j^{(t)}\right) \mid b_j^{(t)}\in\mathcal{U}_t \right\},\: \rho_j^{(t)} = p\left(b_j^{(t)} \mid q,\mathcal{S}_{t-1}\right)
\end{equation}
where $\rho_j^{(t)}$ represents the selection probability for candidate $b_j^{(t)}$ out of $|\mathcal{C}_t|$ candidates. Under this formulation, deterministic construction sequences become conditionally dependent on all prior actions.

These candidates branch the search space into $i\leq |\mathcal{C}_t|$ new partial sequences, one for each selected block $b_j^{(t)}$. To model these diverging paths, let $\mathcal{V}_t$ represent the multi-hypothesis distribution over partial sequences of length $t$. Formally, $\mathcal{V}_t$ expands each partial sequence $\mathcal{S}^{(t-1)}_i$ from the previous time step via its sequence-conditioned candidate distribution $\mathcal{U}_{t,i}$:
\begin{equation}
    \begin{split}
        &\mathcal{V}_t = \left\{ \left( \langle \mathcal{S}^{(t-1)}_i, b_i^{(t)} \rangle,\ p(\mathcal{S}_i^{(t)} \mid q) \right) \mid b_i^{(t)} \sim \mathcal{U}_{t,i} \right\}, \\
        &p(\mathcal{S}^{(t)}_i \mid q) = p\left(b_i^{(t)} \mid \mathcal{S}^{(t-1)}_{i}, q\right) \cdot p\left(\mathcal{S}^{(t-1)}_i \mid q\right)
    \end{split}
\end{equation}
where the sequence likelihood $p(\mathcal{S}^{(t)}_i \mid q)$ updates once a block is chosen.

\subsection{Sequential Monte Carlo Inference}\label{sect:SMC}
Since exact search over the branching multi-hypothesis sequence space $\mathcal{V}_t$ for a prompt $q$ is computationally intractable, we frame build sequence generation as Sequential Monte Carlo search (Alg.~\ref{alg:seq_monte_carlo}). $N$ parallel particles maintain trajectory hypotheses, each querying its sequence-conditioned candidate set $\mathcal{C}_{t,i}$.

\noindent\textbf{Initialization:} Given an initial build sequence $\mathcal{S}_0$, $N$ parallel particle trajectories $\mathbf{X} = (\mathcal{S}_0)_{i=1}^N$ are established with uniform weights $\mathbf{w} = \frac{1}{N}\mathbf{1}_N$ and an empty pool of completed sequences $\mathcal{P} = \emptyset$.

\noindent\textbf{Prediction:} At step $t$, candidate generator function $\mathcal{F}\left(q,\ \mathcal{S}_i^{(t-1)}\right)$ yields candidate set $\mathcal{C}_{t,i}$ for each active particle. We expand each particle to construct uncorrected sequence proposals $\hat{\mathcal{S}}_{i,j}^{(t)} = \left\langle \mathcal{S}_i^{(t-1)}, b_j^{(t)} \right\rangle$ with initial probabilities $\rho_j^{(t)}$.

\begin{algorithm}[t]
\caption{Sequential Monte Carlo inference for build-sequence generation.}
\label{alg:seq_monte_carlo}
\small
\begin{algorithmic}[1]
\REQUIRE Initial sequence $\mathcal{S}_0$, prompt $q$, particle count $N$, and maximum horizon $T$
\ENSURE Selected build sequence $\mathcal{S}^{\star}$

\STATE $\mathbf{X} \gets (\mathcal{S}_0)_{i=1}^{N}, \quad \mathbf{w} \gets \frac{1}{N}\mathbf{1}_N, \quad \mathcal{P} \gets \emptyset$

\FOR{$t = 1, \ldots, T$}
    \STATE $\mathcal{C}_{t,i} \gets \mathcal{F}\left(q,\ \mathcal{S}_i^{(t-1)}\right)$ for $i = 1, \ldots, |\mathbf{X}|$
    \STATE $\hat{\mathcal{S}}_{i,j}^{(t)} \gets \left\langle \mathcal{S}_i^{(t-1)}, b_j^{(t)} \right\rangle$ for $i = 1, \ldots, |\mathbf{X}|, b_j^{(t)} \in \mathcal{C}_{t,i}$

    \STATE $\mathbf{v}_{i,j}^{(t)} \gets \operatorname{MTV}_{xy}\left(\hat{\mathcal{S}}_{i,j}^{(t)}\right)$ for $i = 1, \ldots, |\mathbf{X}|, b_j^{(t)} \in \mathcal{C}_{t,i}$
    \STATE $\gamma_{i,j}^{(t)} \gets \rho_j^{(t)} \exp\!\left(-\left\lVert \mathbf{v}_{i,j}^{(t)} \right\rVert_2\right)$ for $b_j^{(t)} \in \mathcal{C}_{t,i}$
    \STATE $\bar{\mathcal{S}}_{i,j}^{(t)} \gets \left\langle \mathcal{S}_i^{(t-1)}, b_j^{(t)} + \mathbf{v}_{i,j}^{(t)} \right\rangle$ for $b_j^{(t)} \in \mathcal{C}_{t,i}$

    \STATE $\eta_i^{(t)} \gets \sum_{b_j^{(t)} \in \mathcal{C}_{t,i}} \gamma_{i,j}^{(t)}, \quad \mathcal{I}_t \gets \{i : \eta_i^{(t)} > 0\}$
    
    \IF{$\mathcal{I}_t = \emptyset$}
        \STATE \textbf{break}
    \ENDIF
    
    \STATE $\pi_i(j) \gets \gamma_{i,j}^{(t)} \Big/ \eta_i^{(t)}$ for $i \in \mathcal{I}_t, b_j^{(t)} \in \mathcal{C}_{t,i}$
    
    \STATE $\bar{\mathcal{S}}_i^{(t)} \gets \bar{\mathcal{S}}_{i,j^*}^{(t)},\quad j^* \sim \operatorname{Cat}(\pi_i)$ for $i \in \mathcal{I}_t$
    
    
    \STATE $w_i^{(t)} \gets w_i^{(t-1)} \cdot \pi_i(j^*)$ for $i \in \mathcal{I}_t$
    
    \STATE $\mathcal{P} \gets \mathcal{P} \cup \left\{ \left(\bar{\mathcal{S}}_i^{(t)}, w_i^{(t)}\right) : i \in \mathcal{I}_t, \operatorname{Last}\left(\bar{\mathcal{S}}_i^{(t)}\right) = \mathrm{EOS} \right\}$
    
    \STATE $\mathcal{A} \gets \left\{ i \in \mathcal{I}_t : \operatorname{Last}\left(\bar{\mathcal{S}}_i^{(t)}\right) \neq \mathrm{EOS} \right\}$
    
    \IF{$\mathcal{A} = \emptyset$}
        \STATE \textbf{break}
    \ENDIF
    
    \STATE $\bar{w}_i^{(t)} \gets w_i^{(t)} \Big/ \sum_{k \in \mathcal{A}} w_k^{(t)}$ for $i \in \mathcal{A}$
    
    \STATE $\mathbf{X}^{(t)} \gets \operatorname{Resample}_{|\mathcal{A}|}\left( \left\{\bar{\mathcal{S}}_i^{(t)}\right\}_{i \in \mathcal{A}}, \left\{\bar{w}_i^{(t)}\right\}_{i \in \mathcal{A}} \right)$
    
    \STATE $\mathbf{X} \gets \mathbf{X}^{(t)}, \quad \mathbf{w} \gets \frac{1}{|\mathcal{A}|}\mathbf{1}_{|\mathcal{A}|}$
\ENDFOR

\IF{$\mathcal{P} = \emptyset$}
    \STATE $\mathcal{P} \gets \left\{\left(\mathcal{S}_i, w_i\right)\right\}_{i=1}^{|\mathbf{X}|}$
\ENDIF

\RETURN $\operatorname*{arg\,max}_{(\mathcal{S}, w) \in \mathcal{P}} \left( \operatorname{Length}(\mathcal{S}), w \right)$

\end{algorithmic}
\end{algorithm}

\noindent\textbf{Measurement and Repair:} Proposed candidates may violate physical non-overlap constraints. To preserve spatial feasibility while avoiding costly full stability checks during online expansion, collisions in $\hat{\mathcal{S}}_{i,j}^{(t)}$ are repaired via a 2D constrained Minimum Translation Vector $\operatorname{MTV}_{xy}\left(\hat{\mathcal{S}}_{i,j}^{(t)}\right)$, yielding displacement $\mathbf{v}_{i,j}^{(t)}$ and repaired sequence $\bar{\mathcal{S}}_{i,j}^{(t)} = \left\langle \mathcal{S}_i^{(t-1)}, b_j^{(t)} + \mathbf{v}_{i,j}^{(t)} \right\rangle$. To account for this displacement, candidates receive collision-adjusted penalty scores:
\begin{equation}
    \gamma_{i,j}^{(t)} = \rho_j^{(t)} \exp\!\left(-\left\lVert \mathbf{v}_{i,j}^{(t)} \right\rVert_2\right)
\end{equation}
where candidates requiring larger spatial corrections receive higher exponential penalties, and unresolvable collisions are assigned a score of zero.

\noindent\textbf{Reweight and Resample:} For active particles $\mathcal{I}_t$ with positive mass $\eta_i^{(t)} = \sum_{b_j^{(t)} \in \mathcal{C}_{t,i}} \gamma_{i,j}^{(t)} > 0$, candidate scores are normalized to yield placement probabilities:
\begin{equation}
    \pi_i(j) = \frac{\gamma_{i,j}^{(t)}}{\eta_i^{(t)}}
\end{equation}
Each active particle samples a block index $j^* \sim \operatorname{Cat}(\pi_i)$ to obtain the new predicted sequence $\bar{\mathcal{S}}_i^{(t)} = \bar{\mathcal{S}}_{i,j^*}^{(t)}$ and the updated weight $w_i^{(t)} = w_i^{(t-1)} \cdot \pi_i(j^*)$, which models the recursive sequence likelihood $p(\mathcal{S}_i^{(t)} \mid q)$. Completed sequences ending in $\mathrm{EOS}$ are harvested into pool $\mathcal{P}$. Remaining active particles $i \in \mathcal{A}$ are normalized to $\bar{w}_i^{(t)}$ and importance-resampled to form the next ensemble $\mathbf{X}^{(t)}$.

\noindent\textbf{Selection:} Generation terminates at horizon $T$ or when $\mathcal{A} = \emptyset$. The final sequence $\mathcal{S}^\star = \operatorname*{arg\,max}_{(\mathcal{S}, w) \in \mathcal{P}} \left( \operatorname{Length}(\mathcal{S}), w \right)$ is selected by maximizing sequence length, breaking ties with cumulative particle weight $w$.

\section{Multimodal Task-Specific Architecture}
\begin{figure*}[hb]
    \centering
    \includegraphics[width=1.0\textwidth]{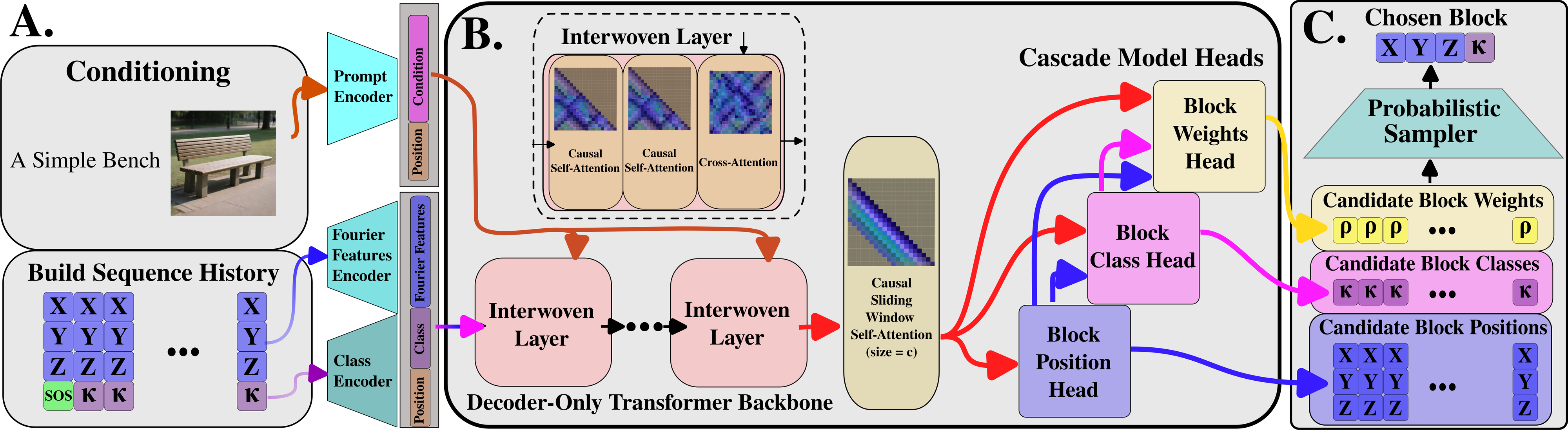}

    \caption{\textbf{A.} Model input: a partial build sequence of blocks and a conditioning prompt (e.g., text, images or some combination of both). \textbf{B.} Our proposed architecture. We utilize several interwoven layers stacked in series as our transformer backbone. The output of the final interwoven layer is fed to a windowed attention layer that ties relevant local candidate-level features to each other before being passed to three cascading model heads that predict the candidate block positions, classes and likelihoods. \textbf{C.} We select the next block through probabilistic Sampling using either Top-$K$ or SMC.}
    \label{fig:architecture}
\vspace{-0.5cm}
\end{figure*}
Parameterizing the generator function $\mathcal{F}\left(q,\ \mathcal{S}_i^{(t-1)}\right)$ introduced in Section~\ref{sect:SMC}, ScaffoldM3C uses an autoregressive transformer architecture (Figure~\ref{fig:architecture}) to map a partial build sequence to a distribution over next-block proposals, conditioned on a user-specified text or image prompt. During inference, these proposals drive Sequential Monte Carlo search to generate feasible build sequences, selecting the longest valid trajectory upon termination.

To process input sequences, we separately encode block positions, block shape classes, and the conditioning prompt. Since Fourier Features enhance spatial understanding, attention to detail, and training efficiency~\cite{tancik2020fourier}, we lift the block positions into a higher dimensional latent space via a learned high-frequency Fourier mapping, $E_{\text{FF}} = [\sin(2\pi Bx),\ \cos(2\pi Bx)]$, where $B$ denotes the frequency matrix. Block classes, belonging to a fixed-size lexicon, are mapped to learned categorical token embeddings. We then concatenate the transformed position and class embeddings into a single input vector. Conditioning prompts are encoded separately using a 4-bit quantized Gemma 3 model~\cite{gemma_2025} to extract rich, pretrained semantic vectors. We capture sequence order using both learned absolute positional embeddings~\cite{devlin2019bert}, which are concatenated with input vectors, and relative positional embeddings (RoPE)~\cite{su2024roformer}, which are applied before each self-attention layer~\cite{grattafiori2024llama}. Combining both methods yields superior structural quality compared to using either alone.

The model backbone integrates separate build and conditioning sequences through $M_{\text{L}}$ blocks of interwoven causal and cross-attention layers. Causal attention captures global context from prior block placements, while cross-attention injects conditioning information to align generated sequences with user objectives.  After extracting useful global features, a windowed attention layer, with window size equal to the candidate pool size, correlates local features to associate missing structural elements with potential placement locations. Inputs to each layer are normalized using RMSNorm~\cite{zhang2019root} alongside QK normalization~\cite{henry2020query} to prevent instability and attention weight collapse.

Due to the strong interdependency between position and class selection, we predict the candidate set 
\begin{equation}
\left\{ \left( (\hat{P}^{(t)}_j, \hat{\kappa}^{(t)}_j), \hat{\rho}_j^{(t)} \right) \right\}_{j=1}^{|\mathcal{C}_t|} = \mathcal{F}_\theta\left(q,\ \mathcal{S}_i^{(t-1)}\right)
\end{equation}
at time $t$ using three cascaded fully connected heads. First, the position head outputs $|\mathcal{C}_t|$ candidate positions from learned features. Next, the class head processes these positions alongside attention features to output $|\mathcal{C}_t|$ categorical distributions over $K$ shapes. Finally, the likelihood head combines latent features with predicted positions and classes to yield a $|\mathcal{C}_t|$-dimensional vector, representing the placement likelihood for each candidate at step $t$.

During training, we frame the candidate proposal step as a simplified Mixture of Experts (MoE), where each candidate generated by the model heads acts as an ``expert'' proposing a potential position, class, and confidence score. The loss function is a weighted sum of task prediction terms ($\mathcal{L}_{\text{task}}=\mathcal{L}_{\text{pos}}+\mathcal{L}_{\text{class}}$), routing terms ($\mathcal{L}_{\text{router}}=\mathcal{L}_{\rho}$), expert utilization balance ($\mathcal{L}_{\text{aux}}$)~\cite{fedus2022switch}, and router stability ($\mathcal{L}_{\text{z-loss}}$)~\cite{zoph2022st}. Beyond standard MoE objectives, we introduce a cosine similarity loss term, $\mathcal{L}_{\text{diversity}}$, which enforces candidate diversity by penalizing identical position-class predictions. The total loss is defined as
\begin{equation}
    \mathcal{L} = \mathcal{L}_{\text{task}} + \mathcal{L}_{\text{router}} + \mathcal{L}_{\text{aux}} + \mathcal{L}_{\text{z-loss}} + \mathcal{L}_{\text{diversity}}
\end{equation}
See the Appendix~\ref{appx:loss_functs} for further implementation details.

\section{Multimodal Stability-Aware Dataset Generation}
\label{sec:dataset}

We build our training corpus upon the StableText2Brick dataset~\cite{pun2025generating}, which provides five textual descriptions and an ordered build sequence for each structure. 
Although these sequences satisfy StableText2Brick's stability criteria, they do not satisfy our stability constraints for every intermediate assembly.
\begin{figure}[ht]
    \centering
    \includegraphics[width=1\linewidth, trim={0 0 0 0}, clip]{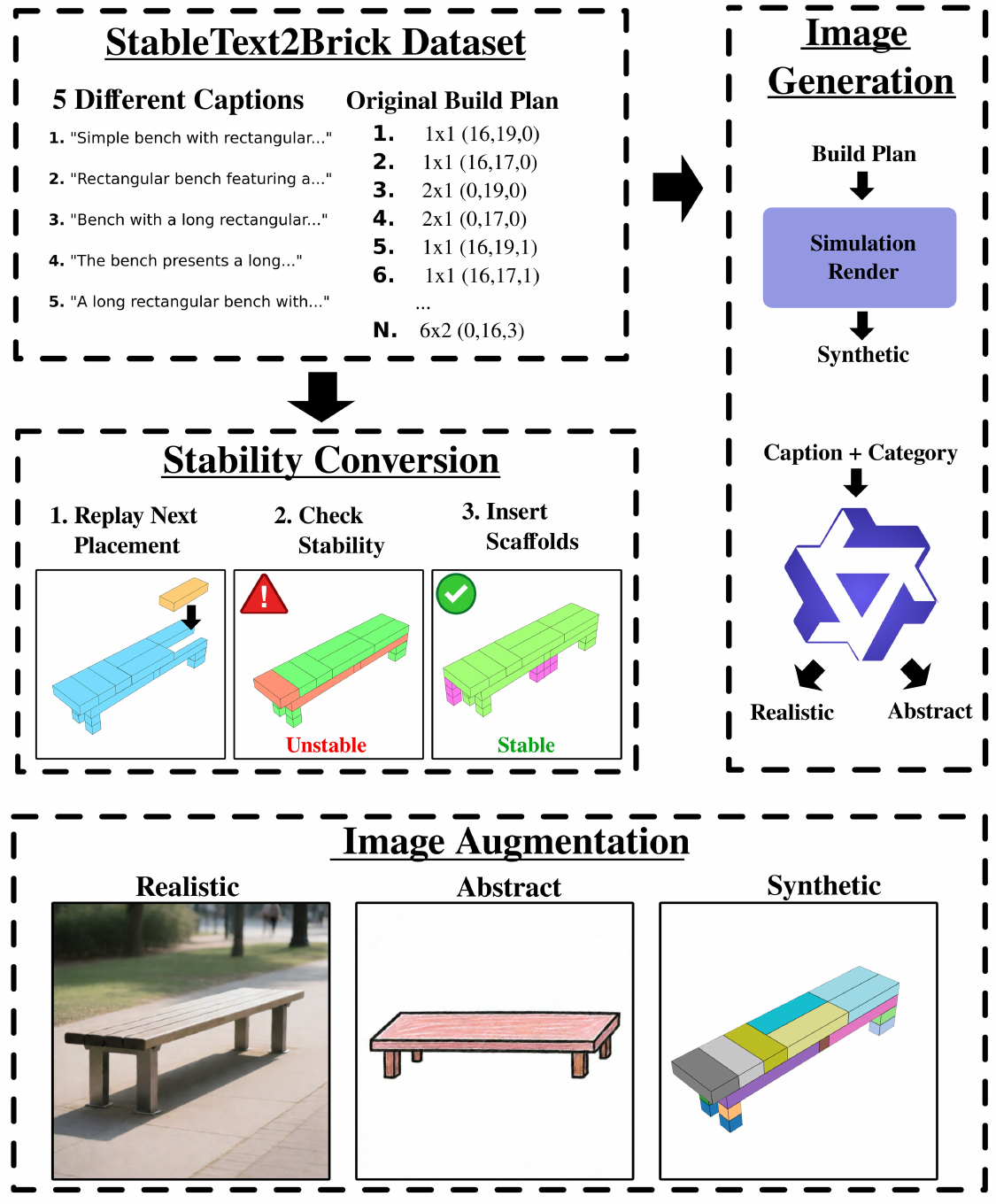}
    \caption{Data generation pipeline extending the StableText2Brick dataset. Build sequences are updated with scaffold blocks to satisfy our stability constraints. Images are generated either by simulating the construction and rendering the final structure specified by a given build plan or by providing a text-based guidance prompt to a Qwen text-to-image model.}
    \label{fig:dataset_gen}
\vspace{-5pt}
\end{figure}
We therefore replay each sequence $\mathcal{S}_{\text{orig}}$, evaluate the stability of every block placement, and insert the minimum number of scaffold blocks required to support it. This yields a stability-aware sequence $\mathcal{S}$ 
with fully stable intermediate and final assemblies while preserving structure geometry and target block placement order, aligning training supervision with Section~\ref{sec:problem_formulation}.

We further extend the dataset with three complementary visual conditions: \emph{synthetic}, \emph{realistic}, and \emph{abstract}. Synthetic images are obtained by executing $\mathcal{S}_{\text{orig.}}$ in simulation and rendering the completed structure from a sampled camera viewpoint, preserving exact block geometry. For the realistic and abstract conditions, we combine the structure category with one of its textual descriptions using style-specific guidance templates and pass the resulting prompts to a Qwen text-to-image model. The realistic condition depicts the target object in a natural scene, whereas the abstract condition provides a simplified artistic interpretation. These complementary domains diversify the conditioning space and reduce reliance on a single rendering style (Figure~\ref{fig:dataset_gen}).

For training, each construction sequence is expanded into 13 condition--sequence pairs: five text-only, three image-only, and five multimodal. Each multimodal condition pairs one of the five captions with a visual condition sampled from the three image domains. All conditions share the same stability-aware target sequence $\mathcal{S}$, enabling unified training across text, image, or combined inputs. A pretrained Gemma-based encoder~\cite{gemma_2025} maps each conditioning input $q$ to a latent token sequence $\mathbf{E}_q$, yielding the final training pair $(\mathbf{E}_q,\mathcal{S})$. This pipeline transforms the original text--structure dataset into a stability-aware multimodal corpus, paving the way for learning construction tasks where scaffolding is required to stabilize complex geometries, such as overhangs.

\section{Experiments}

We benchmark our method against BrickGPT~\cite{pun2025generating}, a state-of-the-art 1B-parameter LLaMA-3.2-Instruct model fine-tuned to generate assembly sequences from text instructions. By default, BrickGPT employs an expensive post-processing pipeline that combines rejection sampling with checkpoint reversion to guarantee feasible, stable structures. Since this strategy is independent of the model and inference heuristic, any probabilistic approach can benefit from it. Therefore, to ensure a fair architectural comparison, we evaluate two baseline variants: one with post-processing entirely disabled to mirror our single forward-pass setup, and another with rollback disabled but step-wise block rejection enabled to compare against our test-time collision-checking framework. Additionally, we fine-tune a scaffold-aware variant, BrickGPT-Scaffold, which utilizes rejection sampling and is evaluated under the same metrics.

Beyond comparing against various BrickGPT variants, we evaluate the utility of Sequential Monte Carlo against Top-$K$ sampling, a popular inference heuristic for probabilistic token selection, when processing our model's output to generate build sequences.

We evaluate each method in simulation across 479 text and 288 image prompts taken from the test set to ensure statistical significance. Results are presented in Table~\ref{tab:baseline_table}, where '---' indicates an incompatible conditioning modality.
Furthermore, we perform an ablation study on a subset of 24 text-based prompts to explore the trade-offs of Sequential Monte Carlo inference as population size increases, visually analyze candidate block belief states as the sequence grows, and examine how varying candidate and class distribution temperatures affect assembly plan quality. Finally, to demonstrate real-world feasibility, we autonomously construct several physical structures using an xArm Lite robotic manipulator.

\subsection{Evaluation Metrics}
We group our metrics into three categories: stability, feasibility, and semantic alignment. We further distinguish between metrics that capture overall trends and those that provide per-sample statistical averages. Let $\mathcal{D} = \left\{\mathcal{S}_1, \mathcal{S}_2, \dots, \mathcal{S}_d \right\}$ represent the test set consisting of $d$ generated structures, where each structure $\mathcal{S}_k$ consists of $N_k$ blocks.

\noindent\textbf{Stability Metrics:} When evaluating structure stability, we distinguish between intermediate stability, $s_{\text{inter}}$, and final stability, $s_{\text{final}}$. We define the intermediate stability score as the average per-block stability of a structure during construction, evaluated by measuring the stability of each new block with respect to only previously placed blocks:
\begin{equation}
s_{\text{inter}}=\frac{1}{d}\sum_{k=1}^{d}\frac{1}{N_k}\sum^{N_k}_{t=1}s\left(b^{(t)}_{k}\mid \mathcal{S}^{(t-1)}_k\right)
\end{equation}
where $s\left(\cdot\right)\in\left\{0,1\right\}$ evaluates the stability of block $b^{(t)}_k$. In contrast, inspired by the definition in BrickGPT~\cite{pun2025generating}, we define the final stability score as the average per-block stability of a structure assuming all other blocks in the structure have already been placed:
\begin{equation}
s_{\text{final}}=\frac{1}{d}\sum_{k=1}^{d}\frac{1}{N_k}\sum^{N_k}_{t=1}s\left(b^{(t)}_k\mid \mathcal{S}_{k}\setminus\left\{b^{(t)}_k\right\}\right)
\end{equation}
Beyond tracking how average structure stability evolves over time, $s_{\text{inter}}$ and $s_{\text{final}}$ allow us to estimate the rate at which a method places a block out of order. Following BrickGPT~\cite{pun2025generating}, we define the overall stability rate, $s_{\text{overall}}$, as the percentage of constructed structures in the test set that are fully sequentially stable:
\begin{equation}
s_{\text{overall}}=\frac{100}{d}\sum_{k=1}^{d}\mathbb{I}_{\text{stable}}\left(\mathcal{S}_k\right)
\end{equation}
where $\mathbb{I}_{\text{stable}}(\mathcal{S}_k)=\prod^{N_k}_{t=1}s\left(b^{(t)}_k\mid \mathcal{S}^{(t-1)}_{k}\right)$ is an indicator function specifying whether a structure is fully stable.

\begin{table*}[ht!]
  \centering
  \caption{Average quantitative results without rollbacks across $479$ text and $288$ image test prompts. Arrows indicate direction of improvement, while $\approx$ specifies the desired target. Performance is highlighted as \textbf{\textcolor{darkgreen}{best}}, \textbf{\textcolor{darkblue}{second best}}, and \textbf{\textcolor{red}{worst}}.}
    \scalebox{0.7}{
    \input{Tables/unified_table}
    \label{tab:baseline_table}
  }
  \vspace{-0.1cm}
\end{table*}

\noindent\textbf{Feasibility Metrics:} Besides stability, a critical measure of structural feasibility is the number of collisions that would occur if a build plan is enacted. By definition, structures with fewer collisions are more feasible. Accordingly, we adopt the feasibility metrics defined in BrickGPT~\cite{pun2025generating}. We measure the average per-block collision, $e_{\text{collision}}$, as the average number of voxels occupied by two or more blocks for a given structure:
\begin{equation}
e_{\text{collision}}=\frac{1}{d}\sum_{k=1}^{d}\frac{1}{N_k}\sum^{N_k}_{m=1}\sum^{N_k}_{n=1,\ m\neq n} f_{\text{occ}}\left(b^{(m)}_k,b^{(n)}_k\right)
\end{equation}
where $f_{\text{occ}}(b^{(m)}_k, b^{(n)}_k) = |V(b^{(m)}_k) \cap V(b^{(n)}_k)|$ measures the number of intersecting voxels between blocks $b^{(m)}_k$ and $b^{(n)}_k$, with $V(b_k)$ representing the set of voxels occupied by a block $b_k$ in structure $\mathcal{S}_k$. We also measure the overall collision-free rate, $e_{\text{rate}}$, observed as the percentage of constructed structures from the test set that are collision-free:
\begin{equation}
e_{\text{rate}}=\frac{100}{d}\sum_{k=1}^{d}\mathbb{I}_{\text{free}}\left(\mathcal{S}_k \right)
\end{equation}
where $\mathbb{I}_{\text{free}}(\mathcal{S}_k)=\mathbb{I}\left( \sum_{m=1}^{N_k}\sum_{n=1,\ m\neq n}^{N_k} f_{\text{occ}}\left(b^{(m)}_k,b^{(n)}_k\right) = 0\right)$ is an indicator function for detecting if a construction plan is collision-free. Finally, we determine the overall feasibility rate, $e_{\text{feasible}}$, by evaluating whether each structure $\mathcal{S}_k$ is both fully stable and collision-free, averaging over the test set:
\begin{equation}
    e_{\text{feasible}} = \frac{100}{d}\sum_{k=1}^{d}\left(\mathbb{I}_{\text{stable}}(\mathcal{S}_k)\cdot \mathbb{I}_{\text{free}}(\mathcal{S}_k)\right)
\end{equation}
This establishes a strict metric acknowledging that true physical feasibility requires both complete structural stability and collision-free steps.

\noindent\textbf{Semantic Alignment Metrics:} To quantify prompt alignment, we measure the prompt-render similarity score, $a_{\text{sim}}$, using CLIP for text prompts and DINOv2 for image prompts against a rendered view of the constructed structure. We obtain the ground-truth alignment score, $a_{\text{gt}}$, by measuring the DINOv2 similarity between rendered views of predicted and ground-truth structures.
To evaluate generalization beyond the training set, we compute two 3D vertex-level Chamfer distances: the minimum Chamfer distance, $c_{\text{min}}$, to any training structure, where higher values indicate greater novelty, and the semantic Chamfer distance, $c_{\text{sem}}$, to the ground-truth structure with the highest prompt similarity to its own instruction to assess prompt generation flexibility. 

\begin{figure*}[ht!]
    \centering
    \includegraphics[width=1.\linewidth]{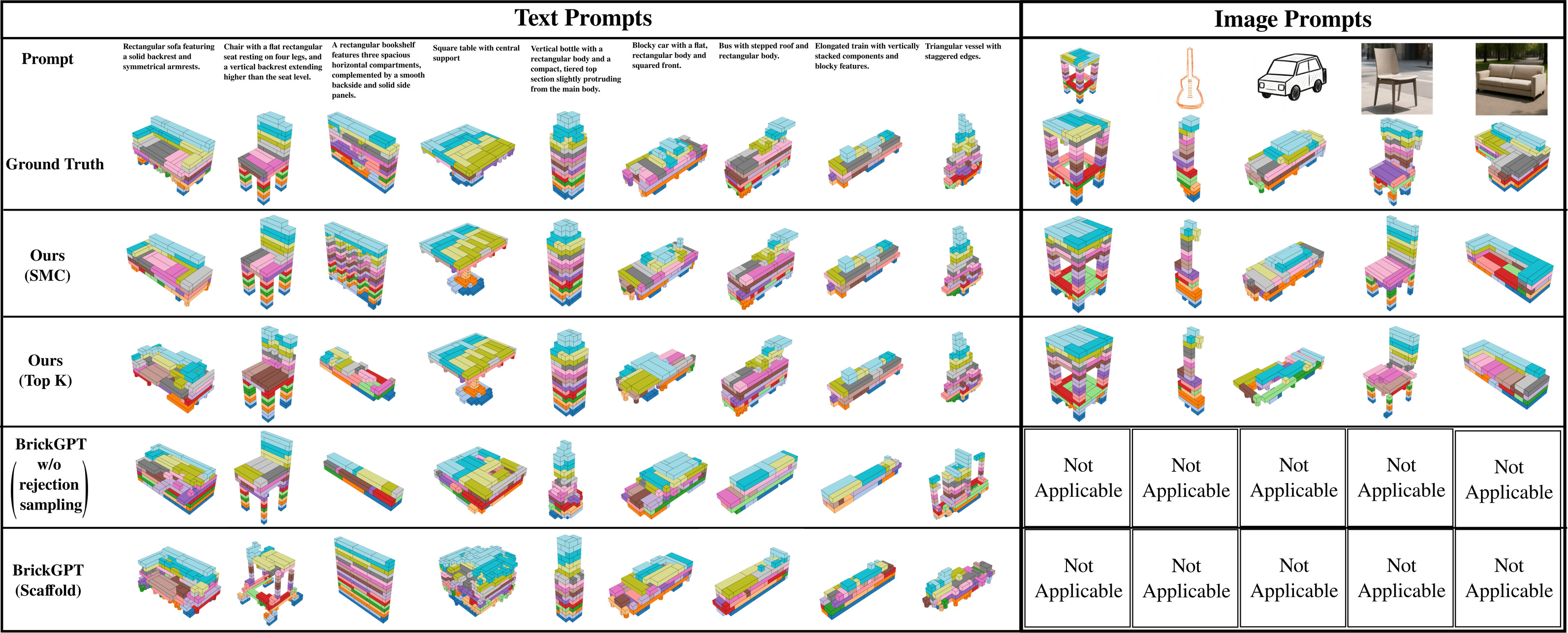}
    \caption{Qualitative results of our method using both text and image prompts taken from the test dataset. We compare against our baseline using text prompts. Our baseline cannot interpret image prompts, therefore we leave ``Not Applicable'' for its results.}
    \label{fig:results_monte_carlo}
\end{figure*}

\subsection{Structure Quality and Feasibility}
\noindent\textbf{Visual Quality:} As demonstrated in Figure~\ref{fig:results_monte_carlo}, our framework generates visually appealing structures that align well with the provided text or image prompt, without requiring rejection sampling or checkpoint reversion to repair inference errors. Furthermore, structures constructed using Sequential Monte Carlo (SMC) are visually comparable to those created by the BrickGPT variants and produce fewer artifacts than those seen in Top-$K$ sampling. This gap stems from the limitations of Top-$K$ sampling, which is restricted to candidates conditioned on a single past build sequence and easily becomes trapped in local minima during inference. In contrast, SMC maintains a belief distribution over diverse, feasible sequence candidates, enabling the framework to explore alternative assembly plans and recover from suboptimal block placements made early in the generation process.

\noindent\textbf{Feasibility:} Besides comparable visual quality, our approach eliminates block intersections entirely, boosting the collision-free rate from the $90.19\%$ achieved by BrickGPT-Scaffold to $100\%$ (see Table~\ref{tab:baseline_table}). Moreover, unlike rejection sampling methods, which require hyperparameter tuning, introduce sequential bottlenecks, and cannot ensure collision-free solutions under tight assembly constraints, our multi-block formulation allows us to parallelize collision evaluation, repair, and candidate likelihood updates, delivering higher feasibility at minimal inference cost.

\noindent\textbf{Stability:} Both standard BrickGPT variants are incapable of placing scaffold blocks and must rely solely on careful structural placement. While BrickGPT-Scaffold can generate scaffolding, it fails to utilize it effectively, often placing multiple scaffold blocks adjacent to one another rather than distributing them to enforce stability. Conversely, our approach distributes scaffold blocks far more effectively across the structure (see Figure~\ref{fig:results_monte_carlo}), resulting in higher overall stability. These qualitative observations are reinforced by quantitative block counts compared against the ground truth (Table~\ref{tab:baseline_table}). Standard BrickGPT with rejection sampling closely matches the underlying structure distribution, but cannot place scaffolding. BrickGPT-Scaffold incorporates scaffold awareness, but drastically overgenerates blocks, producing approximately 195 blocks and 42 scaffold blocks. In contrast, our SMC-based approach places 127 total blocks and 29 scaffold blocks, consistent with the expected structural scale and scaffolding distribution.

These scaffolding dynamics directly impact the measured structural stability. Under our stability metric, our SMC-based framework outperforms basic BrickGPT baselines and Top-$K$ sampling in overall stability. While BrickGPT-Scaffold achieves higher per-block stability, this is an artifact of its reliance on inherently stable scaffold tokens ($1.45\times$ more scaffold blocks than our approach), which inflates its score. In contrast, our Top-$K$ and SMC methods achieve $8.56\%$ and $15.24\%$ higher overall stability than BrickGPT-Scaffold, respectively. Transitioning from Top-$K$ to SMC yields a $6\%$ relative gain, proving SMC's ability to correct early placement errors via parallel hypothesis tracking.

Finally, across all methods, we observe completed structures exhibiting higher stability than during their intermediate states, signaling occasional out-of-order placements (e.g., placing a block before its supports). This occurs because models rely purely on learned spatial priors. Integrating lightweight, online stability checks during SMC reweighting would prune unsafe paths without rollbacks, which we reserve for future work.

\subsection{Model Semantic Alignment and Generalization}
Our model is highly flexible, capable of responding to diverse instruction modalities, including text, images, or a mixture of both. We observe strong semantic alignment across both of our inference methods (see Table~\ref{tab:baseline_table}). When evaluated via text cosine similarity ($a_\text{sim}$), our approach matches standard BrickGPT and BrickGPT-Scaffold. We observe an identical trend when comparing the visual similarity of generated structures to the ground truth using DINOv2 embeddings ($a_\text{gt}$), where SMC achieves a higher mean similarity ($\num{0.8448}$) than our BrickGPT baselines ($\num{0.8234}$) and Top-$K$ sampling ($\num{0.8145}$). These trends are mirrored under pure image instructions as well.

Beyond strong semantic alignment, our architecture demonstrates equivalent generalization capabilities despite using only $225$M parameters, which is more than $4\times$ smaller than the $1$B-parameter BrickGPT. While the minimum Chamfer distance ($c_\text{min}$) is nominally highest for Top-$K$ ($\num{1.1479}$ compared to $\num{0.9006}$ for the baseline), this metric is artificially inflated by Top-$K$'s tendency to build smaller, simpler structures due to early terminations. This conclusion is substantiated by Top-$K$'s lower average block placement ($90$ blocks placed compared to the ground-truth average of $111$) and weaker semantic alignment. In contrast, SMC matches BrickGPT in structural scale, semantic alignment, and minimum Chamfer distance ($\num{0.8056}$). This confirms that SMC achieves strong generalization without compromising structural complexity or visual quality. We observe a similar trend in semantic generalization: SMC ($\num{4.1993}$) maintains a minimum semantic Chamfer distance ($c_\text{sem}$) highly comparable to the baseline ($\num{4.5665}$), while avoiding the structural degradation that artificially inflates Top-$K$'s score ($\num{5.5036}$).

\subsection{Inference Speed}
A major advantage of our approach is its computational efficiency. As shown in 
Table~\ref{tab:baseline_table}, our SMC framework is at least $5\times$ faster than standard BrickGPT and up to $36\times$ faster than BrickGPT-Scaffold. Similarly, our approach using Top-$K$ is $3\times$ faster than our SMC framework and $20\times$ to $111\times$ faster than BrickGPT. Although Top-$K$ generally produces coarser assembly plans, its rapid inference speed makes it a cost-effective drop-in replacement for BrickGPT in the rejection sampling and checkpoint reversion pipeline proposed by \cite{pun2025generating}, potentially allowing online robot planning. 

Conversely, our SMC framework allows users to explicitly trade inference speed for structural quality by adjusting the particle population size and tuning candidate and class likelihood temperatures. Decreasing the population size allows SMC to approach Top-$K$ speeds at the cost of hypothesis diversity, whereas increasing the population size allocates additional compute to explore alternative assembly decisions, a capability particularly beneficial for building complex structures. Similarly, lower temperatures reduce exploration by concentrating on high-likelihood candidate blocks, whereas higher temperatures encourage exploration, helping escape deadlocks caused by tight assembly constraints. We examine these SMC trade-offs in detail in Sections~\ref{abl:smc_belief} and~\ref{abl:candates}.

\subsection{Robot Hardware Experiments}
Since our model is computationally lightweight, we combine our SMC framework with rejection sampling to find feasible structures and then demonstrate real world feasibility by autonomously constructing the resulting structures using a 6-DOF XFactory robotic arm. The structures are assembled using 3D-printed large LEGO-style blocks, where the block dimensions are as specified in~\cite{pun2025generating}.  Furthermore, before deploying the generated assembly plan, the stand-in scaffold block tokens are physically instantiated. We replace each token with the requisite number of singleton scaffold blocks, stacking them from the ground or underlying support block up to the position specified by the token. 

\begin{figure}[h]
    \centering
    \includegraphics[width=1.\linewidth]{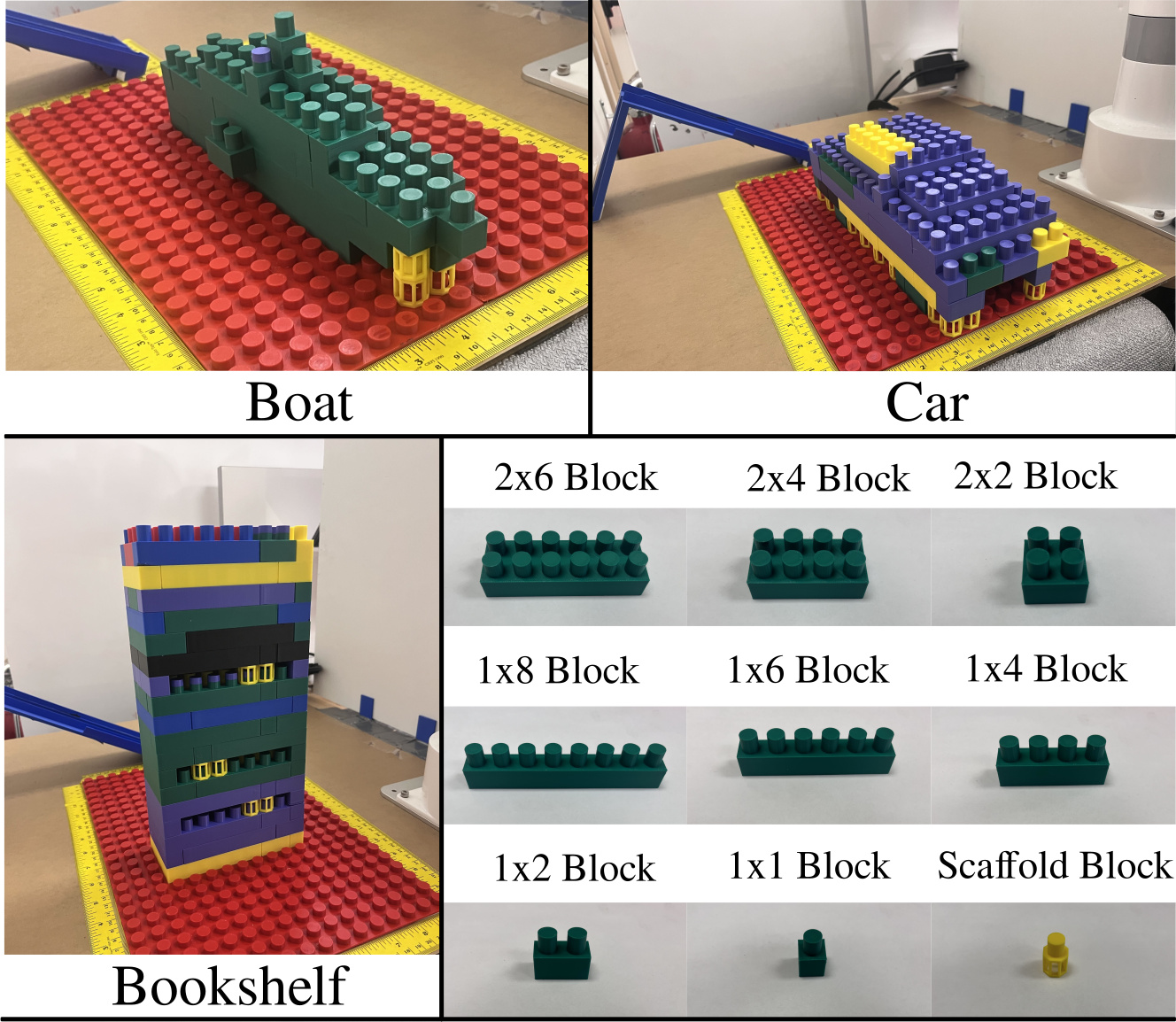}
    \caption{Hardware experiments using a 6-DOF XFactory robotic arm to build three different structures. During assembly, nine different blocks were used. Overhanging blocks that would normally be unstable are supported by scaffolding.}
    \label{fig:hardware}
\end{figure}

The construction plans are executed using open-loop position control coupled with a compliant torque-feedback loop, allowing the blocks to be slotted safely and accurately into their desired locations. The results of these physical experiments are shown in Figure~\ref{fig:hardware}.

As illustrated in Figure~\ref{fig:hardware}, our framework effectively utilizes scaffolds to stabilize overhangs, such as the prow of the ship, the bumpers of the car, and the shelves of the bookshelf. In addition, the structures generated are complex, using 45, 86, and 89 blocks for the boat, car, and bookshelf, respectively. 

\section{Ablations}

\subsection{Sequential Monte Carlo Belief State}\label{abl:smc_belief}
\begin{figure*}[ht!]
    \centering
    \includegraphics[width=1.\linewidth]{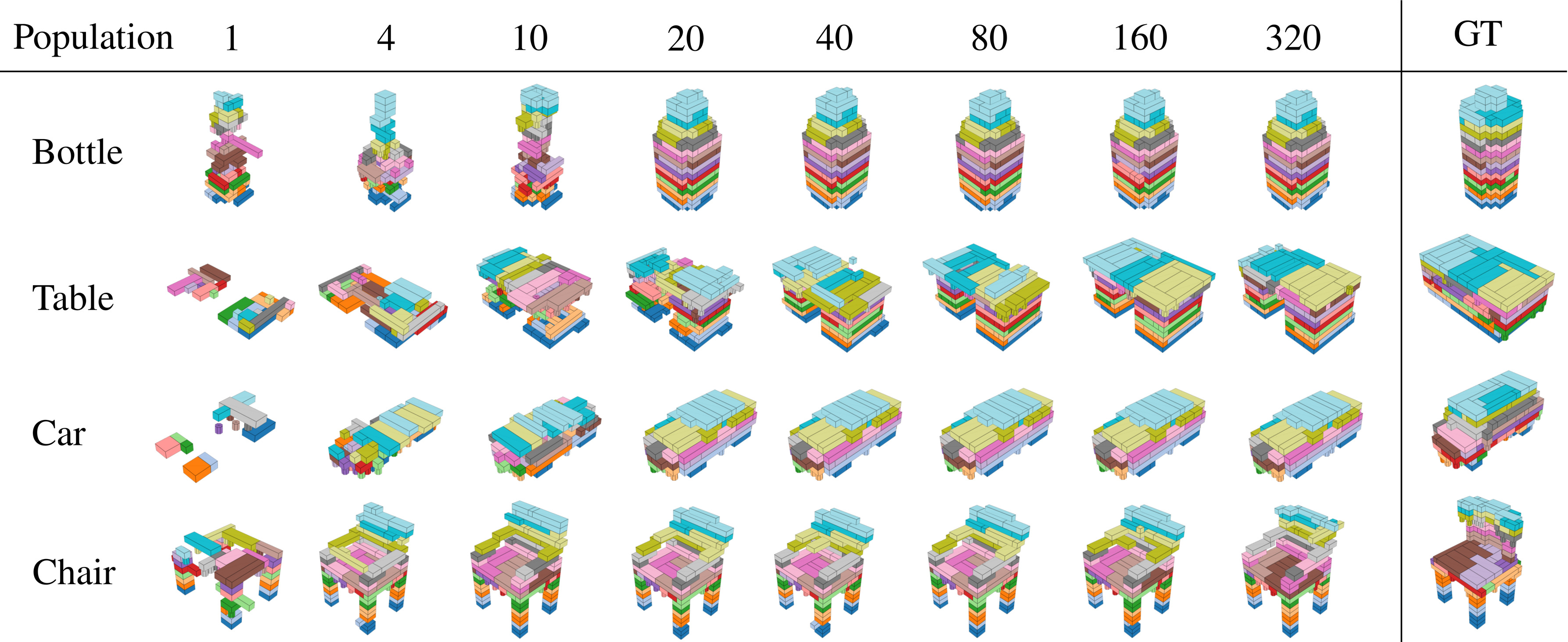}
    \caption{Effect of Sequential Monte Carlo (SMC) population size on generated structures. Each row shows results for a different target object as the number of candidate build sequences increases from 1 to 320, with the ground-truth (GT) structure shown on the right. Larger populations generally yield more complete and coherent structures that better match the target geometry, with diminishing improvements beyond 160 sequences.}
    \label{fig:pop_ablation}
\end{figure*}

\begin{figure}[h]
    \centering
    \includegraphics[width=1\linewidth]{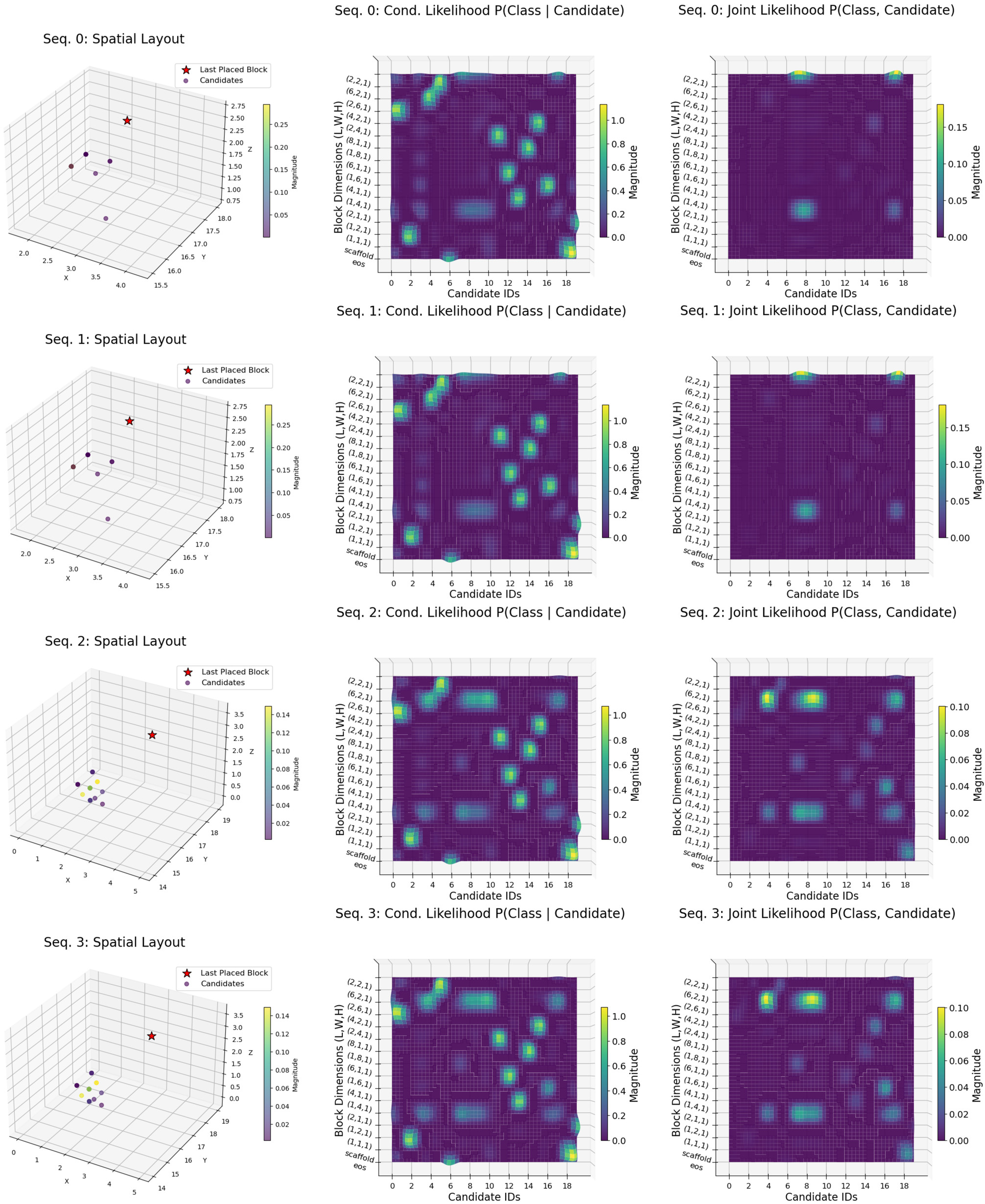}
    \caption{Belief state of Sequential Monte Carlo after 12 steps when asked to generate a table. A population size of 4 was used. Each belief state considers a separate and independent set of candidate blocks, conditioned on previously placed blocks of the associated assembly sequence. Multiple particles allow for alternative block placement decisions to be considered simultaneously, allowing the model to potentially escape local minima.}
    \label{fig:smc_distribution}
\end{figure}

To examine the effect that the build sequence population has on the resulting structural quality, we ablate population sizes used by SMC for 1, 4, 10, 20, 40, 80, 160, and 320 different sequences. Figure~\ref{fig:smc_distribution} clearly demonstrates the utility of considering multiple sequences simultaneously, as after 6 steps with just four different sequences considered at once, our framework has already begun examining distinct build sequences, each with its own set of candidate positions and likelihoods. Furthermore, from Table~\ref{tab:pop_ablation} and Figure~\ref{fig:pop_ablation}, we observe that there is an upward trend for stability and structural quality as population size increases, with diminishing returns beyond 160 sequences. As expected, this improvement comes at the cost of increased inference time where the best trade-off between build quality and inference time occurs at a population size between 20 and 40.
\begin{table}[ht!]
  \centering
  \caption{Ablation Study on Sequential Monte Carlo Population Size using 24 different test samples.}
    \label{tab:pop_ablation}
    \input{Tables/pop_ablation_table}
\end{table}

\subsection{Candidate Versus Class Likelihood}\label{abl:candates}
Although increasing the population size in general results in improved structures, we observe that it is still possible for the model to terminate structures early or generate invalid structures even at higher population sizes. 
\begin{figure}[h]
    \centering
    \includegraphics[width=1.\linewidth]{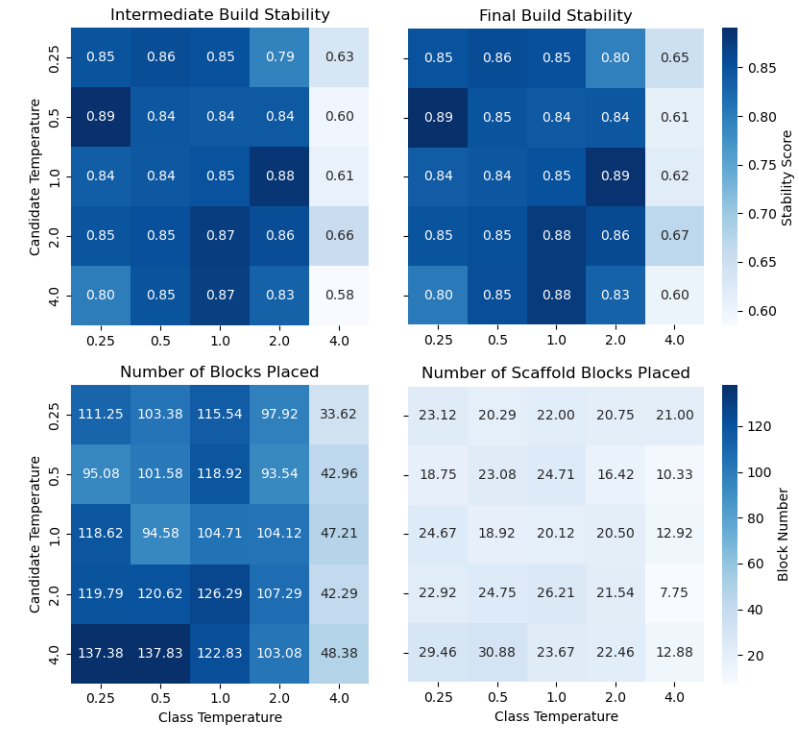}
    \caption{Heatmaps of intermediate stability, final stability, standard block count, and scaffold block count under varying candidate and class distribution temperatures (evaluated across 24 test samples). Structural stability is particularly sensitive to the class distribution temperature, with higher temperatures potentially favoring implausible block-type selections and disrupting the assembly sequence. In contrast, increasing the candidate distribution temperature generally improves stability and increases both standard and scaffold block counts, suggesting that broader exploration of lower-probability placement locations allows SMC to identify more successful build sequences.}
    \label{fig:heatmap}
\end{figure}
These failures are likely due to implausible build sequences surviving the pruning phase, creating a runaway effect where the model begins placing blocks at random upon entering an out-of-distribution (OOD) region. 

We can reduce this effect by tuning the candidate block and class distributions using a temperature hyperparameter for each distribution. To observe how temperature affects each distribution, we performed a hyperparameter sweep with temperatures of 0.25, 0.5, 1.0, 2.0, and 4.0 (see Figure~\ref{fig:heatmap}). As shown in the heatmaps, structural stability is very sensitive to changes in class distribution, with stability quickly dropping if the distribution is altered. We also observe a similar trend in the number of standard and scaffold blocks, which indicates that the model is more likely to enter an OOD regime at higher class temperatures. This finding is supported by an increasing difference between intermediate and final stability, which suggests more blocks are being placed out of order. In comparison, we observe that stability, block count, and scaffold count generally increase as the candidate distribution temperature increases. This is likely due to SMC being permitted to consider lower-probability placement locations, enabling broader exploration of the assembly space. Conversely, increasing the class distribution temperature only affects the choice of block type for a given position, where the set of valid options is often dictated by blocks already placed.

\section{Limitations}
Our framework has some limitations. 
First, automatically generated prompts and reference images can be ambiguous, occasionally assigning similar conditions to visually distinct structures; although this may encourage generalization, it can also reduce confidence during block sampling. Second, while scaffold tokens enforce physically stable intermediate configurations, the resulting placement order is not necessarily easy to execute, as previously placed blocks may obstruct workspace clearance and complicate motion planning. In addition, our SMC procedure selects the longest completed sequence, which can bias generation toward larger structures and does not explicitly evaluate stability. Future work could jointly optimize structural and scaffold placement order for robotic accessibility and incorporate explicit stability validation into the generation loop for closed-loop sequence selection.
\subsection{Failure Modes}
\begin{figure}[h]
    \centering
    \includegraphics[width=1.0\linewidth]{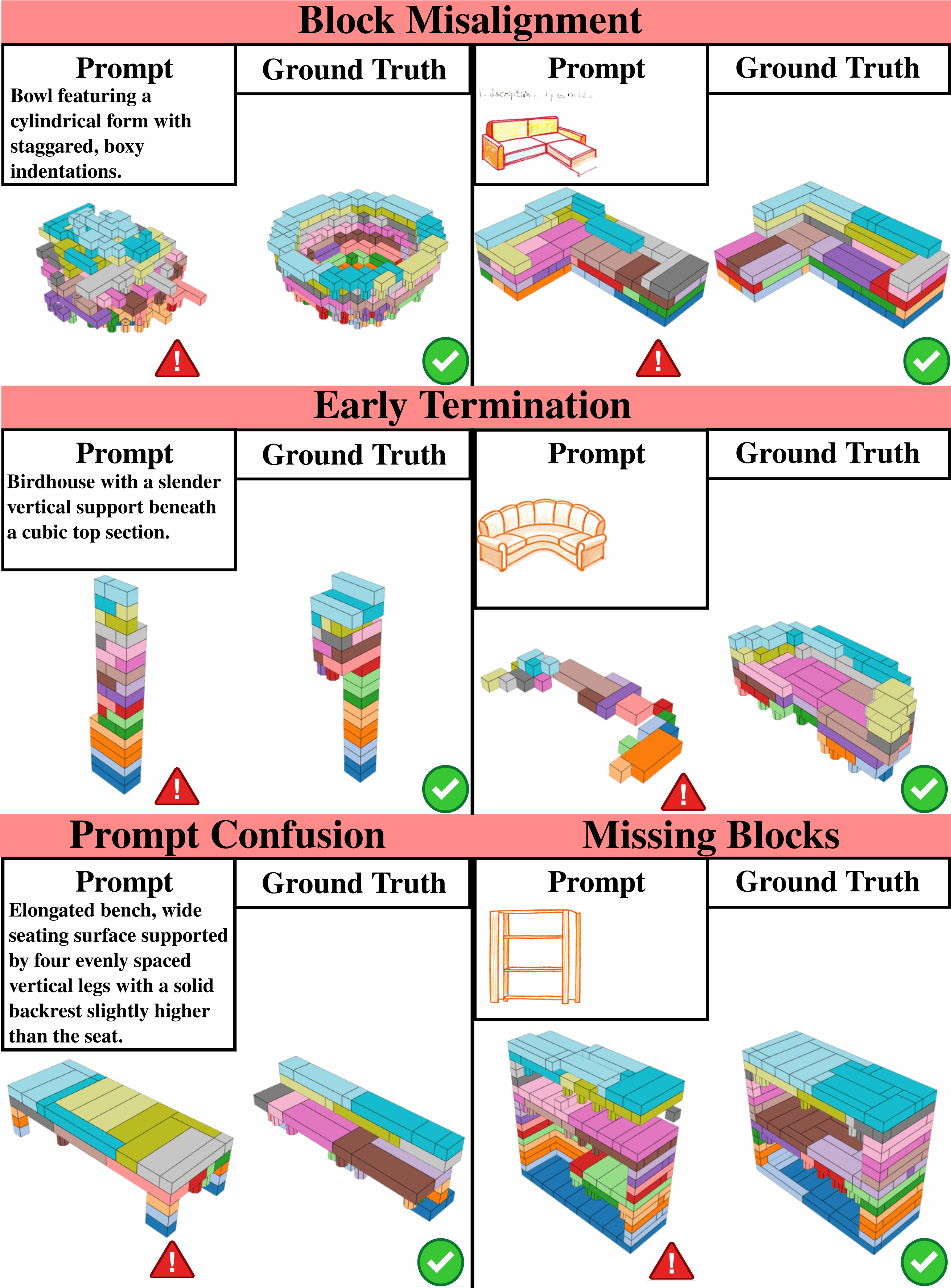}
    \caption{Possible failure modes of sequence generation using Sequential Monte Carlo sampling. \textbf{(1)} Positional errors lead to assembly plan degradation or out-of-distribution (OOD) states. \textbf{(2)} The model misses prompt details, generating a similar but irrelevant structure. \textbf{(3)} The model prematurely outputs an EOS token. \textbf{(4)} Errors in predicted positions result in block shifting.}
    \label{fig:limitations_monte_carlo}
\end{figure}

Our approach exhibits four primary failure modes (Figure~\ref{fig:limitations_monte_carlo}): block misalignment, early termination, prompt confusion, and missing blocks or scaffolding.
The most prevalent issue is block misalignment, where noise shifts predicted candidate positions along cardinal directions. This causes candidate blocks to shift to avoid unexpected collisions or leads to the selection of lower-probability block candidates. The first case can render placement unstable if a candidate fails to meet the conditions in Section~\ref{sec:problem_formulation}, or yield physically stable yet visually and semantically poor structures. Furthermore, it degrades confidence in subsequent steps, compounding downstream placement errors. The second case leads the model into an out-of-distribution (OOD) region, either by selecting an atypical block for a structure category or placing a valid block in an unseen location. This typically degrades candidate confidence and severely compromises the final assembly plan.

Block misalignment frequently triggers early termination, where the model overconfidently predicts a termination token ahead of time (a failure that can also occur from excessively high candidate or class sampling temperatures). Less frequent failure modes include prompt confusion, where the model misinterprets instructions and builds an unrelated structure, and missing blocks or scaffolding. Both often manifest as downstream consequences of an initial block misalignment.

\section{Conclusion}
In this work, we present \textit{ScaffoldM3C}, a lightweight, multimodal, autoregressive model paired with a Sequential Monte Carlo (SMC) framework for stable block-based construction. We formulate assembly as a probabilistic next-block generation task, leveraging candidate proposals to capture action diversity and SMC search to identify stable build sequences, thereby mitigating greedy decoding errors. Unlike existing approaches, our method explicitly integrates scaffolding via auxiliary support tokens, enabling the model to resolve intermediate instability during assembly by placing temporary support structures. In addition, we extend the StableText2Brick dataset~\cite{pun2025generating} with image conditioning prompts and scaffold-stabilized build sequences to support future research in vision-language-construction models.

Our framework requires $4\times$ fewer parameters than competing baselines while delivering a $5\times$ to $20\times$ inference speedup. It generates structures that match baseline visual quality and achieve superior physical stability, without requiring computationally expensive sequence rollbacks. Lastly, we validate that the resulting build plans are physically feasible in both simulated and real-world automated robotic assembly.

\bibliographystyle{IEEEtran}
\bibliography{references} 


\appendices
\section{Loss Function Terms}\label{appx:loss_functs}

We group our loss terms used in our modified MoE loss function into two categories: the task loss and the candidate balance loss terms. The task loss is composed of a weighted sum of terms: MSE for $P_t,$,  Cross Entropy for $\kappa_t$, and point-wise KL Divergence Loss for $\rho_t$, that try to make accurate individual predictions for each expert. Similarly, the balance loss consists of the sum of four weighted terms: An auxiliary loss $L_{\text{aux}}$ that penalizes unequal utilization of candidates, a candidate Z-loss $L_{\text{z, candidate}}$ the guards against training instability for candidate selection, a class Z-loss $L_{\text{z, class}}$ that guards against training instability for class selection, and a cosine similarity diversity term $L_{\text{diversity}}$ that enforces class diversity.

Our task loss is made up of three terms: the Sequence Mean Square Error (SMSE) loss, the Sequence Cross Entropy (SCE) loss, and Pointwise KL Divergence (KL) loss. We compute the SMSE using
\begin{equation}
    L_{\text{pos}}=\frac{1}{N_k}\sum^{N_k}_{t=1}\left(P^{(t)} - \hat{P}^{(t)}_{j^*}\right)^2
\end{equation}
which compares the position of the highest likelihood candidate, $\hat{P}^{(t)}_{j^*}$, with the ground truth block position, $P^{(t)}$, for the position $t$ in the sequence. The SCE loss 
\begin{equation}
    L_{\text{class}}=-\frac{1}{N_k}\sum^{N_k}_{t=1} \kappa^{(t)}\ \text{log}\left(\hat{\kappa}^{(t)}_{j^*}\right)
\end{equation}
compares the class logits of the highest likelihood candidate, $\hat{\kappa}^{(t)}_{j^*}$, with the ground truth block class, $\kappa^{(t)}$ for each block in the sequence. Next, we treat the predictions generated by the probability head as if it were the router in a standard MoE architecture. 
To obtain a useful distribution over the possible candidates, we utilize the KL loss,
\begin{equation}
    L_{\text{router}}=\frac{1}{N_k}\sum^{N_k}_{t=1} \rho^{(t)}_j\left(\text{log}\ \rho^{(t)}_j - \text{log}\ \hat{\rho}^{(t)}_j\right)
\end{equation}
where $\hat{\rho}^{(t)}_j$ is predicted likelihood of a candidate and $\rho^{(t)}_j$ is the target likelihood. Since our training data that does not contain multiple ground truth candidates blocks and their associated placement likelihood for each step in the assembly process, we approximate this distribution by treating the negative distance from each predicted candidate to the same ground truth block as our likelihood weights, $\rho^{(t)}_j=-(L_{MSE,\ t} + L_{CE,\ t})$.

To prevent mode collapse we employ a load-balancing auxiliary loss over the $|\mathcal{C}_t|$ candidate predictions:
\begin{equation}
    L_{\text{aux}} = \left( \frac{K}{k_{\text{top}}} \sum_{c=1}^{K} f_c \cdot \bar{P}_c \right) - 1
\end{equation}
\begin{align}
    \bar{P}_c &= \frac{1}{|\mathcal{C}_t|} \sum_{j=1}^{|\mathcal{C}_t|} P_{j,\kappa}, \\
    f_\kappa &= \frac{1}{|\mathcal{C}_t|} \sum_{j=1}^{|\mathcal{C}_t|} m_{j,\kappa}, \quad m_{j,\kappa} \in \{0, 1\}
\end{align}
where $P_{j,\kappa} = \text{softmax}(w_{j,\kappa})$ denotes the probability of candidate proposal $j$ belonging to block class dimension $\kappa \in \{1, \dots, K\}$.
Here, $m_{j,\kappa}$ is a binary indicator denoting whether class $\kappa$ ranks within the $k_{\text{top}}$ predictions for candidate proposal $j$. The candidate Z-loss
\begin{equation}
    L_{\text{z, candidate}} = \frac{1}{|\mathcal{C}_t|} \sum_{j=1}^{|\mathcal{C}_t|} \left( \log \sum_{\kappa=1}^{K} \exp\left(w_{j,\kappa}\right) \right)^2
\end{equation}
operates on raw router logit weights, $w_{j,\kappa}$ while the class Z-loss
\begin{equation}
    L_{\text{z, class}} = \frac{1}{|\mathcal{C}_t|} \sum_{j=1}^{|\mathcal{C}_t|} \left( \log \sum_{\kappa=1}^{K} \exp\left(z_{j,\kappa}\right) \right)^2
\end{equation}
Operates on predicted class logits $z_{j,\kappa}$ for candidate $j$ and class $\kappa$. To prevent spatially overlapping candidate proposals from predicting redundant class distributions, we penalize pairwise cosine similarity between non-top-$k_{\text{top}}$ pairs within distance $0.1$:
\begin{equation}
\mathcal{L}_{\text{diversity}}^{(t)}
=
\frac{
\sum_{n \neq m}^{|\mathcal{C}_t|}
\left( \hat{\mathbf{p}}_n^\top \hat{\mathbf{p}}_m \right)
\cdot
\mathbb{I}\left(\mathbf{x}_n,\mathbf{x}_m\right)
\cdot
\left(1-T_nT_m\right)
}{
\max\!\left(
1,\;
\sum_{n \neq m}^{|\mathcal{C}_t|}
\mathbb{I}\left(\mathbf{x}_n,\mathbf{x}_m\right)
\cdot
\left(1-T_nT_m\right)
\right)
}.
\label{eq:diversity_loss}
\end{equation}
where $\hat{\mathbf{p}}_n = \frac{\text{softmax}(\mathbf{z}_n)}{\|\text{softmax}(\mathbf{z}_n)\|_2} \in \mathbb{R}^K$, $\mathbf{x}_n \in \mathbb{R}^3$ is the spatial candidate position, $\mathbb{I}\left(\mathbf{x}_n,\mathbf{x}_m\right)=\mathbb{I}\!\left(
\left\| \operatorname{sg}(\mathbf{x}_n)
-
\operatorname{sg}(\mathbf{x}_m) \right\|_2 < 0.1
\right)$ is an indicator function that determines if candidate n and m should be penalized, $\text{sg}(\cdot)$ is the stop-gradient operator, and $T_n \in \{0,1\}$ indicates top-$k_{\text{top}}$ selection.

\section{Training and Hyperparameters}\label{appx:hyper_para}
We train a mix-modal model using distributed data parallel training on two A6000 GPUs for 1500 epochs using a Pytorch Lightning framework and we used a drop out value of 0.1 and learning rate of $5\times10^{-4}$. The model has 225M trainable parameters with a total hidden dimension size of 1024, where the hidden dimension is split into 512, 384 and 128 for the block position, block class and sequence positional embedding respectively. We configured the model to predict 20 different candidates during each forward pass. When using Sequential Monte Carlo for inference, we maintain a population of 20 different feasible build sequences and use a temperature value of 2 and 1 for the candidate and class likelihoods respectively. This choice of temperature was selected to balance between exploration and exploitation of candidates presented by the model.  When using Top-$K$ sampling, we instead specified that 80 different block position and type combinations would be considered each forward pass and set the temperature of both the candidate and class likelihoods to be 0.1 to prevent the model from being too uncertain in its selection of possible candidates.

\end{document}

%% file: Tables/unified_table.tex
{\setlength{\tabcolsep}{4pt}   
\begin{tabular}{@{}l l ccccccccccccc@{}}
\toprule
    & &  \multicolumn{7}{c}{\textbf{Per Structure}} & \multicolumn{6}{c}{\textbf{Across Structures}}\\
    \cmidrule(lr){3-9} \cmidrule(lr){10-15}
& Method & $e_\text{collision} \downarrow$ & $s_\text{inter}\uparrow$ &$s_\text{final}\uparrow$ & $a_\text{sim}\uparrow$ & $a_\text{gt}\uparrow$ & $c_\text{min}\uparrow$ & $c_\text{sem}\uparrow$  & blocks $\approx$ & scaffolds $\approx$ & $s_\text{overall}\uparrow$ & $e_\text{rate}\uparrow$ & $e_\text{feasible}\uparrow$ & Inference Time (s)$\downarrow$\\
\midrule
\multirow{7}{*}{\rotatebox[origin=c]{90}{Text}}
& \textcolor{gray!100}{Ground truth} & \textcolor{gray!100}{\ms{0.000}{0.000}} & \textcolor{gray!100}{\ms{1.000}{0.000}} & \textcolor{gray!100}{\ms{1.000}{0.000}} & \textcolor{gray!100}{\ms{1.000}{0.000}} & \textcolor{gray!100}{\ms{1.000}{0.000}} & \textcolor{gray!100}{N/A} & \textcolor{gray!100}{N/A} & \textcolor{gray!100}{\ms{111.0271}{70.0497}} & \textcolor{gray!100}{\ms{28.0313}{22.3448}} & \textcolor{gray!100}{$\num{100}\ \%$} & \textcolor{gray!100}{$\num{100}\ \%$} & \textcolor{gray!100}{$\num{100}\ \%$} & \textcolor{gray!100}{N/A}\\ 

& BrickGPT w/o rej. & \cellcolor{lightred}{\ms{7.0566}{10.1357}} & \cellcolor{lightred}{\ms{0.7638}{0.1107}} & \cellcolor{lightred}{\ms{0.7644}{0.1101}} & \ms{0.2692}{0.0288}  & \ms{0.8234}{0.0775} & \ms{0.9006}{0.9280} & \ms{4.5665}{5.5986} & \cellcolor{lightcyan}{\ms{106.1500}{72.1203}} & \ms{0.0000}{0.0000} & \cellcolor{lightred}{$\num{2.7254}\ \%$} & \cellcolor{lightred}{$\num{28.0922}\ \%$} & \cellcolor{lightred}{$\num{2.3061}\ \%$} & \ms{11.4646}{9.7937}\\

& BrickGPT w/ rej. & \ms{0.0000}{0.0000} & \ms{0.7567}{0.1118} & \ms{0.7568}{0.1117} & \ms{0.2689}{0.0285}  & \ms{0.8203}{0.0813} & \cellcolor{lightcyan}{\ms{0.9175}{1.1091}} & \cellcolor{lightcyan}{\ms{4.6517}{5.6934}} & \cellcolor{lightgreen}{\ms{110.9102}{83.2579}} & \ms{0.0000}{0.0000} & $\num{3.1315}\ \%$ & $\num{100.0000}\ \%$ & $\num{3.1315}\ \%$ & \ms{13.8635}{16.8635}\\

& BrickGPT-Scaffold & \ms{0.4280}{2.4044} & \cellcolor{lightgreen}{\ms{0.9183}{0.0557}} & \cellcolor{lightgreen}{\ms{0.9188}{0.0552}} & \ms{0.2671}{0.0285}  & \cellcolor{lightcyan}{\ms{0.8326}{0.0728}} & \ms{0.8231}{0.6533} & \ms{4.2404}{5.2198} & \cellcolor{lightred}{\ms{195.1023}{158.9575}} & \cellcolor{lightred}{\ms{42.4029}{43.6773}} & $\num{11.4823}\ \%$ & $\num{90.1879}\ \%$ & $\num{11.2735}\ \%$ & \cellcolor{lightred}{\ms{64.5053}{215.9353}}\\

& ours (Top-$K$)& \cellcolor{lightcyan}{\ms{0.0000}{0.0000}} & \ms{0.8149}{0.1619} & \ms{0.8192}{0.1580} & \ms{0.2695}{0.0269}  & \ms{0.8145}{0.0885} & \cellcolor{lightgreen}{\ms{1.1479}{1.1855}} & \cellcolor{lightgreen}{\ms{5.5036}{6.2879}} & \ms{89.5908}{55.2851} & \cellcolor{lightcyan}{\ms{21.3132}{17.8939}} & \cellcolor{lightcyan}{$\num{20.0418}\ \%$} & \cellcolor{lightcyan}{$\num{100.0000}\ \%$} & \cellcolor{lightcyan}{$\num{20.0418}\ \%$} & \cellcolor{lightgreen}{\ms{0.5827}{0.3574}}\\

& ours (SMC) & \cellcolor{lightgreen}{\ms{0.0000}{0.0000}} & \cellcolor{lightcyan}{\ms{0.8330}{0.1507}} & \cellcolor{lightcyan}{\ms{0.8356}{0.1484}} & \ms{0.2708}{0.0272}  & \cellcolor{lightgreen}{\ms{0.8448}{0.0803}} & \cellcolor{lightred}{\ms{0.8056}{0.7153}} & \cellcolor{lightred}{\ms{4.1993}{5.2676}} & \ms{126.9395}{81.1712} & \cellcolor{lightgreen}{\ms{29.1879}{26.8442}} & \cellcolor{lightgreen}{$\num{26.7223}\ \%$} & \cellcolor{lightgreen}{$\num{100.0000}\ \%$} & \cellcolor{lightgreen}{$\num{26.7223}\ \%$} & \cellcolor{lightcyan}{\ms{1.7862}{2.0332}}\\
\midrule
\multirow{6}{*}{\rotatebox[origin=c]{90}{Image}}
& \textcolor{gray!100}{Ground truth} & \textcolor{gray!100}{\ms{0.000}{0.000}} & \textcolor{gray!100}{\ms{1.000}{0.000}} & \textcolor{gray!100}{\ms{1.000}{0.000}} & \textcolor{gray!100}{\ms{1.000}{0.000}} & \textcolor{gray!100}{\ms{1.000}{0.000}} & \textcolor{gray!100}{N/A} & \textcolor{gray!100}{N/A}  & \textcolor{gray!100}{\ms{110.9062}{66.7700}} & \textcolor{gray!100}{\ms{28.7222}{22.4924}} & \textcolor{gray!100}{$\num{100}\ \%$} & \textcolor{gray!100}{$\num{100}\ \%$} & \textcolor{gray!100}{$\num{100}\ \%$} & \textcolor{gray!100}{N/A} \\

& BrickGPT w/o rej. & \textemdash & \textemdash & \textemdash & \textemdash & \textemdash & \textemdash & \textemdash &\textemdash &\textemdash &\textemdash &\textemdash &\textemdash &\textemdash\\

& BrickGPT w/ rej. & \textemdash & \textemdash & \textemdash & \textemdash & \textemdash & \textemdash & \textemdash &\textemdash &\textemdash &\textemdash &\textemdash &\textemdash &\textemdash\\

& BrickGPT-Scaffold  & \textemdash & \textemdash & \textemdash & \textemdash & \textemdash & \textemdash & \textemdash &\textemdash &\textemdash &\textemdash &\textemdash &\textemdash &\textemdash\\

& ours (Top-$K$)& \ms{0.0000}{0.0000} & \ms{0.8095}{0.1509} & \ms{0.8146}{0.1473} & \ms{0.3686}{0.3063}  & \ms{0.8198}{0.0795} & \ms{1.1051}{0.9628} & \ms{4.7825}{5.4780} & \ms{94.9549}{55.5147} & \ms{23.4931}{18.4577} & $\num{18.4028}\ \%$ & $\num{100.0000}\ \%$ & $\num{18.4028}\ \%$ & \ms{0.6158}{0.3551}\\

& ours (SMC) & \ms{0.0000}{0.0000} & \ms{0.8231}{0.1499} & \ms{0.8257}{0.1475} & \ms{0.3774}{0.3129}  & \ms{0.8448}{0.0740} & \ms{0.9101}{0.8337} & \ms{4.2496}{5.2786} & \ms{128.1771}{78.3683} & \ms{30.2188}{23.7548} & $\num{20.4861}\ \%$ & $\num{100.0000}\ \%$ & $\num{20.4861}\ \%$ & \ms{2.1261}{2.1053}\\
\bottomrule
\end{tabular}
}

%% file: Tables/pop_ablation_table.tex
{\setlength{\tabcolsep}{4pt}
\begin{tabular}{@{}cccc@{}} 
    \toprule
    \textbf{Population} & \textbf{$s_\text{overall}\uparrow$} & \textbf{$s_\text{final}\uparrow$} & \textbf{Inference Time (s) $\downarrow$}\\
    \midrule
    1 & $\num{20.8333}\ \%$ & \ms{0.7977}{0.1890} & \ms{0.4216}{0.2961}  \\
    4 & $\num{20.8333}\ \%$ & \ms{0.8120}{0.1645} & \ms{0.5560}{0.3301} \\
    10 & $\num{20.8333}\ \%$ & \ms{0.8531}{0.1351} & \ms{0.8866}{0.5078} \\
    20 & $\num{25.0000}\ \%$ & \ms{0.8773}{0.1215} & \ms{1.5963}{1.2019} \\
    40 & $\num{20.8333}\ \%$ & \ms{0.8785}{0.1051} & \ms{2.6385}{2.2327} \\
    80 & $\num{33.3333}\ \%$ & \ms{0.9104}{0.0924} & \ms{4.9324}{4.7820} \\
    160 & $\num{29.1667}\ \%$ & \ms{0.9268}{0.0799} & \ms{11.0897}{10.1484} \\
    320 & $\num{33.3333}\ \%$ & \ms{0.9119}{0.1165} & \ms{22.8945}{18.5414} \\

    \bottomrule
\end{tabular}
}